\documentclass[12pt]{article}
\usepackage[margin=1in]{geometry}
\usepackage{setspace}
\usepackage{amsfonts}       
\usepackage{amsthm}     
\usepackage{booktabs}
\usepackage{float}
\usepackage{graphicx}
\usepackage{hyperref}
\usepackage{mathtools}
\usepackage{microtype}
\usepackage{natbib}
\usepackage{subfig}

\newtheorem{property}{Property}
\newtheorem{proposition}{Proposition}
\newtheorem{example}{Example}

\newcommand{\field}[1]{\mathbb{#1}}
\def \E{\field{E}}
\def \F{\field{F}}
\def \P{\field{P}}

\def\cA{\mathcal{A}}                                
                                
\def\cE{\mathcal{E}}
\def\cF{\mathcal{F}}
\def\cG{\mathcal{G}}
\def\cH{\mathcal{H}}

\def\cM{\mathcal{M}}
\def\cN{\mathcal{N}}
\def\cP{\mathcal{P}}
\def\cR{\mathcal{R}}
\def\cS{\mathcal{S}}
\def\cT{\mathcal{T}}

\def\cV{\mathcal{V}}
\def\cZ{\mathcal{Z}}

\DeclareMathOperator*{\argmax}{argmax}

\begin{document}
{\setstretch{1.0}
\begin{center}
\LARGE
Coordinated Exploration in \\ Concurrent Reinforcement Learning
\end{center}
\begin{center}
\normalsize
\vspace*{6pt}
\large
\begin{minipage}{0.49\textwidth}
	\centering
	Maria Dimakopoulou \\
	Stanford University \\
	\texttt{madima@stanford.edu}
\end{minipage}
\begin{minipage}{0.49\textwidth}
	\centering
	Benjamin Van Roy \\
	Stanford University \\
	\texttt{bvr@stanford.edu}
\end{minipage}
\normalsize
\vspace*{18pt}
\end{center}
}

\begin{abstract}
We consider a team of reinforcement learning agents that concurrently learn to operate in a common environment.
We identify three properties -- {\it adaptivity}, {\it commitment}, and {\it diversity} -- which are necessary for efficient coordinated exploration 
and demonstrate that straightforward extensions to single-agent optimistic and posterior sampling approaches fail to satisfy them.
As an alternative, we propose {\it seed sampling}, which extends posterior sampling in a manner that meets these requirements.
Simulation results investigate how per-agent regret decreases as the number of agents grows, establishing substantial advantages
of seed sampling over alternative exploration schemes.
\end{abstract}

\section{Introduction} \label{intro}

The field of reinforcement learning treats the design of agents that operate in uncertain environments and learn over time to make increasingly effective decisions.  In such settings, an agent must balance between accumulating near-term rewards and probing to gather data from which it can learn to improve longer-term performance.  A substantial literature, starting with \citep{Kearns2002}, has developed reinforcement learning algorithms that address this trade-off in a provably efficient manner.

Until recently, most provably efficient exploration algorithms (e.g., \citep{Jaksch2010}) have been based on upper-confidence-bound approaches.  Over the past few years, new approaches that build on and extend PSRL (posterior sampling for reinforcement learning) \citep{Strens00} have proved advantageous in terms of statistical efficiency \citep{Osband2013,osband2017onoptimistic,osband2017posterior,osband2014model,osband2014near}.

PSRL operates in a simple and intuitive manner.  An agent learns over episodes of interaction with an uncertain environment, modeled as a Markov decision process (MDP).  The agent is uncertain about the transition probabilities and rewards of the MDP, and refines estimates as data is gathered.  At the start of each episode, the agent samples an MDP from its current posterior distribution.  This sample can be thought of as a random statistically plausible model of the environment given the agent's initial beliefs and data gathered up to that time.  The agent then makes decisions over the episode as though the environment is accurately modeled by the sampled MDP.

In concurrent reinforcement learning \citep{silver2013concurrent,pazis2013pac,guo2015concurrent,pazis2016pac}, multiple agents interact with the same unknown environment, share data with one another, and learn in parallel.  One might consider two straightforward extensions of PSRL to the multi-agent setting.  In one, at the start of each episode, each agent samples an independent MDP from the current posterior, which is conditioned on all data accumulated by all agents.  Then, over the course of the episode, each agent follows the decision policy that optimizes its sampled MDP.  The problem with this approach is that each agent does not benefit over the duration of the episode from the potentially vast quantities of data gathered by his peers.  An alternative -- which we will refer to as {\it Thompson resampling} -- would be to have each agent independently sample a new MDP at the start of each time period within the episode, as done in \citep{kim2017thompson}.  The new sample would be from a posterior distribution additionally conditioned on data gathered by all agents up to the time.  However, as discussed in \citep{russo2017tstutorial}, this naive extension is disruptive to the agents' ability to explore the environment thoroughly.  In particular, an agent may have to apply a coordinated sequence of actions over multiple time periods in order to adequately probe the environment.  When MDPs are resampled independently over time periods, agents are taken off course.

The two naive approaches we have discussed highlight potentially conflicting needs to adapt to new information and to maintain the intent with which an agent started exploring the environment.  Efficient coordinated exploration calls for striking the right balance.  In this paper, we present a variation of PSRL -- {\it seed sampling} -- that accomplish	es this.

To focus on the issue of coordinated exploration, we consider a single-episode reinforcement learning problem in which multiple agents operate, making decisions and progressing asynchronously.  In this context, we study the rate at which per-agent regret vanishes as the number of agents increases.  Through this lens, we demonstrate that seed sampling coordinates exploration in an efficient manner and can dramatically outperform other approaches to concurrent reinforcement learning.  In particular, the rate at which regret decays appears to be robust across problems, while for each alternative, there are problems where regret decays at a far slower rate.

\citet{guo2015concurrent}, \citet{pazis2013pac}, and \citet{pazis2016pac} have proposed and studied
UCB exploration schemes for concurrent reinforcement learning.
Advantages of PSRL over UCB in single-agent contexts by themselves motivate extension to concurrent reinforcement learning.
A possibly more important motivating factor, however, is that, as our results demonstrate, UCB approaches sometimes do not coordinate
exploration in an effective manner. 
The issue is that UCB approaches are deterministic, and as such, they do not diversify agent behaviors to effectively
divide and conquer when there are multiple facets of the environment to explore.

A broad range of applications calls for concurrent reinforcement learning.  Many examples can be found in web services,
where each user can be served by an agent that shares data with and learns from the experiences of other agents.
Through coordinated exploration, the agents can efficiently learn to better serve the population of users.
The control of autonomous vehicles presents another important context.
Here, each agent manages a single vehicle, and again, the agents learn from each other as data is gathered. 
The goal could be to optimize a combination of metrics, such as fuel consumption, safety, and satisfaction of transportation objectives. 
Exploratory actions play an important role, and structured diversity of experience may greatly accelerate learning.
The seed sampling algorithms we propose in this paper aim to structure such exploration in a systematic and robust manner.

\section{Problem Formulation} \label{problem}

Consider a time-homogeneous, single-episode MDP, which is identified by $\cM = (\cS, \cA, \cR, \cP, \rho, H)$, where $\cS$ is the finite state space, $\cA$ is the finite action space, $\cR$ is the reward model, $\cP$ is the transition model, $\rho$ is the initial state distribution and $H$ is the horizon. 

Consider $K$ agents, who explore and learn to operate in parallel in this common environment. Each $k$th agent begins at state $s_{k, 0}$ and takes an action at arrival times $t_{k,1}, t_{k,2}, \dots, t_{k, H}$ of an independent Poisson process with rate $\lambda = 1$. At time $t_{k, m}$, the agent takes action $a_{k,m}$, transitions from state $s_{k, m-1}$ to state $s_{k, m}$ and observes reward $r_{k, m}$.
The agents are uncertain about the transition structure $\cP$ and/or the reward structure $\cR$, over which they share common priors. 
There is clear value in sharing data across agents, since there is commonality across what agents aim to learn. 
Agents share information in real time and update their posterior, so that when selecting an action at time $t_{k,m}$, the $k$th agent can base his decision on observations made by all agents prior to that time. 

Denote as $\cT = \left\{0, \dots, \max_{k \in \{1, \dots, K\}} t_{k, H}\right\}$. 
We will define all random variables with respect to a filtered probability space $\left(\Omega, \F, (\F_t)_{t \in \cT}, \P\right)$. 
As a convention, variables indexed by $t$ are $\F_t$-measurable and therefore, variables indexed by $k,m$ are $\F_{t_{k, m}}$-measurable.

The total reward accrued by the agents is 
$\sum_{k=1}^K \sum_{m=1}^{H} r_{k,m}$
and the expected mean regret per agent is defined by 
$$\text{BayesRegret}(K) = \frac{1}{K}\sum_{k=1}^K \E\left[\sum_{m=1}^{H} \left(R^* - r_{k,m}\right)\right]$$
where $R^*$ is the optimal reward.

We now consider some examples that illustrate this problem formulation.

\begin{example}[\textbf{Maximum Reward Path}] \label{random-graph-description}
\normalfont
Consider an undirected graph with vertices $\cV = \{1, \dots, N\}$ and edges $\cE \subseteq \cV \times \cV$. 
The probability of any two vertices being connected is  $p$.
Let $\theta \in \Re_+^{|\cE|}$ be the vector of edge weights. 
We treat $\theta$ as a vector with an understanding that edges are sorted according to a fixed but arbitrary order.  
The state space $\cS$ is the set of vertices $\cV$ and the action space from each vertex $v \in \cV$ is the set of edges incident to $v$. 
When action $(v, u) = e \in \cE$ is taken from state $v$, the agent transitions deterministically to state $u$ and observes reward $r_e$, which is a noisy observation of the weight of edge $e$, such that $\E[r_e | \theta] = \theta_{e}$. 
The $K$ agents are uncertain about the edge weights and share a common $\cN(\mu_0, \Sigma_0)$ prior over $\ln \theta$. 
Denote as $e_{k,m} = (v_{k,m-1}, v_{k,m})$ the $m$th edge of the $k$th agent's path traversed at time $t_{k,m}$.  
For each $m = 1,\ldots, H$ the agent observes a reward $r_{k,m}$, distributed according to $\ln r_{k,m} | \theta \sim \cN(\ln \theta_{e_{k,m}} - \sigma^2/2, \sigma^2)$.
The $K$ agents start from the same vertex $v \in \cV$. 
The objective is, starting from vertex $v$, to traverse the path $(v_0 = v, v_1, \dots, v_H)$ that maximizes $\sum_{m=1}^{H} \theta_{(v_{m-1}, v_m)}$, i.e., to find the maximum reward path from vertex $v$ with exactly $H$ edges. 
\end{example}

\begin{figure}[h]
\centering
\includegraphics[width=0.7\columnwidth]{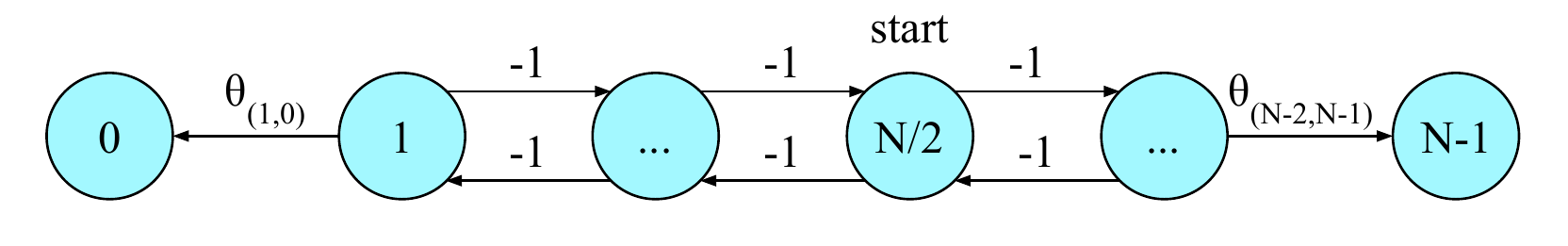}
\caption {Graph of ``Bipolar Chain" example}
\label{bipolar-chain-figure}
\end{figure}

\begin{example}[\textbf{Bipolar Chain}]\label{bipolar-chain-description}
\normalfont
Consider the directed graph of Figure \ref{bipolar-chain-figure}. 
The chain has an even number of vertices, $N$, $\cV = \{0, 1, \dots, N-1\}$.
The endpoints of the chain are absorbing.
The set of edges is $\cE = \{(v, v+1), \forall v = 1, \dots, N-3\} \cup \{(v+1, v), \forall v = 1, \dots, N-3\} \cup (1, 0) \cup (N-2, N-1)$.
The leftmost edge $e_{L} = (1, 0)$ has weight $\theta_{L}$ and the rightmost edge $e_{R} = (N-2, N-1)$ has weight $\theta_{R}$, such that $|\theta_L| = |\theta_R| = N$ and $\theta_R = -\theta_L$.
All other edges $e \in \cE \backslash \{e_L, e_R\}$ have weight $\theta_e = -1$.
The agents do not know whether $\theta_L = N, \theta_R = -N$ or $\theta_L = -N, \theta_R = N$ and they share a common prior that assigns probability $p = 0.5$ to either scenario.
Each one of the $K$ agents starts from vertex $v_{S} = N/2$.
Denote as $e_{k, m}$ the edge traversed at the $m$th step of the $k$th agent's path and $\theta_{k, m}$ the respective weight.
Further, denote as $t_{k, h}$, $1 \leq h \leq H$ the time at which the $k$th agent reaches either endpoint with the $h$th traversal being the last one in the agent's path.
The $k$th agent's objective is to maximize $\sum_{m = 1}^{h} \theta_{k, m}$.
The optimal reward is $R^* = N/2$ if $\theta_L = N, \theta_R = -N$ and $R^* = N/2 + 1$ if $\theta_L = -N, \theta_R = N$ because the leftmost endpoint $v_{L} = 0$ is one vertex further from the start $v_S = N/2$ than the rightmost endpoint $v_{R} = N-1$ and requires the traversal of one more penalizing edge with weight $-1$.
\end{example}

\begin{figure}[h]
\centering
\includegraphics[width=0.5\columnwidth]{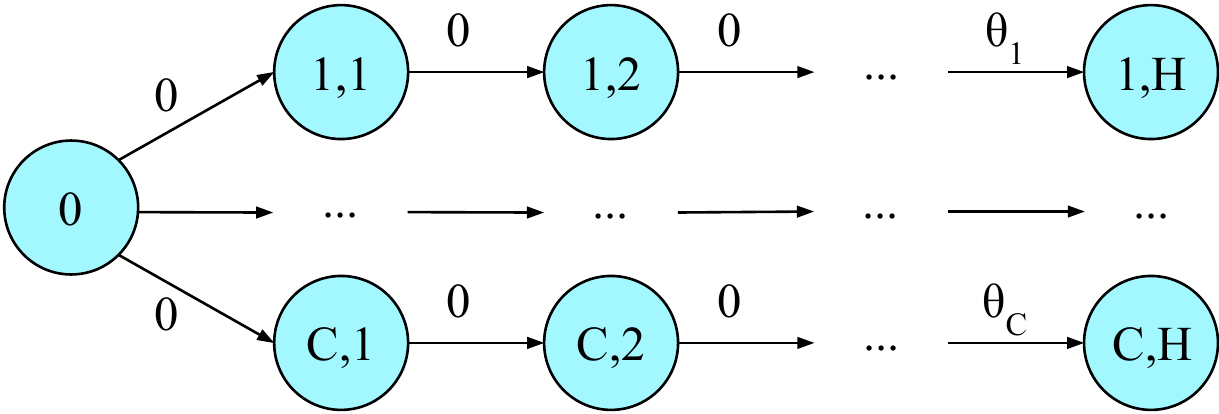}
\caption {Graph of ``Parallel Chains" example}
\label{parallel-chains-figure}
\end{figure}

\begin{example}[\textbf{Parallel Chains}]\label{parallel-chains-description}
\normalfont
Consider the directed graph of Figure \ref{parallel-chains-figure}. 
Starting from vertex $0$, each of the $K$ agents chooses one of the $C$ chains. 
Once a chain is chosen, the agent cannot switch to another chain. 
All the edges of each chain $c$ have zero weights, apart from the edge incoming to the last vertex of the chain, which has weight $\theta_c$.  
Let $\theta \in \Re^{C}$ be the vector of these edge weights for the $C$ chains. 
The $K$ agents are uncertain about $\theta$, over which they share a common $\cN(\mu_0, \Sigma_0)$ prior. 
Denote as $c_k$ the chain chosen by the $k$th agent. 
When traversing the last edge at time $t_{k, H}$, the agent observes reward $r_{k, H}$ distributed according to $r_{k,H} | \theta \sim \cN(\theta_{c_k}, \sigma^2)$.
For all other transitions at times $t_{k, m}$, $m = 1, \dots, H - 1$, the $k$th agent observes reward $r_{k,m} = 0$.
The objective is to choose the chain with the maximum reward.
\end{example}

\section{Algorithms} \label{algorithms}
Three properties are necessary for efficient coordinated exploration in concurrent reinforcement learning:
\begin{property}[\textbf{Adaptivity}] \label{adaptation}
Adapt as data becomes available to make effective use of new information.
\end{property}
\begin{property}[\textbf{Commitment}] \label{intent}
Maintain the intent to carry out probing action sequences that span multiple periods.
\end{property}
\begin{property}[\textbf{Diversity}] \label{diversification}
Divide-and-conquer learning opportunities among agents.
\end{property}
As we discuss in Section \ref{baseline}, straightforward extensions of provably efficient single-agent reinforcement learning algorithms fail to meet these requirements.
In Section \ref{seed sampling}, we introduce the concept of \textit{seed sampling}, which leads to algorithms that simultaneously satisfy these three properties.

All algorithms we consider share some common structure, which we will now describe.
The $K$ concurrent agents share a prior distribution $\cF_0$ of the MDP $\cM$. 
Denote by $\cF_t$ the posterior distribution,  given the history of observations $\cH_{t-1}$ available up to time $t$. 
At each time $t_{k, m}$, the agent generates an MDP $\cM_{k,m}$, computes the optimal policy $\pi_{k,m}$ for $\cM_{k,m}$, takes a single action $a_{k,m} = \pi_{k,m}(s_{k,m-1})$, transitions to state $s_{k,m}$ and observes reward $r_{k,m}$. 
The observation $\left(s_{k,m-1}, a_{k,m}, s_{k,m}, r_{k,m}\right)$ is used to update the shared posterior distribution of $\cM$. 
Therefore, at time $t_{k', m'} > t_{k, m}$, the $k'$th agent can use the knowledge gained from this observation in order to take his $m'$th action. 
The key difference between the studied algorithms is how each $k$th agent forms his MDP $\cM_{k,m}$ at time $t_{k, m}$.

\subsection{Baseline Algorithms} \label{baseline}
First, we discuss the straight-forward adaptation of provably efficient single-agent reinforcement learning algorithms to the concurrent reinforcement learning setting. 
However, neither of these baselines achieve coordinated exploration in concurrent reinforcement learning, either because the agents, when adapting to new information, do not maintain the level of intent required to ensure thorough exploration or because the agents do not diversify their exploratory effort in a manner that mutually benefits their common learning. 

\subsubsection{Thompson Resampling} \label{rpsrl}
At time $t_{k, m}$, the $k$th agent samples MDP $\cM_{k, m}$ from the posterior $\cF_{t_{k, m}}$.
If at time $t_{k, m}$ of the $m$th action of the $k$th agent and at time $t_{k', m'}$ of the $m'$th action of the $k'$th agent the posterior is the same, $\cF_{t_{k,m}} \equiv \cF_{t_{k',m'}}$, the $k$th agent and the $k'$th agent will form a different MDP.
Therefore, the agents will diversify their exploration efforts.
However, resampling an MDP independently at each time period may break the agent's commitment to a sequence of actions that extend over multiple time periods. 
This commitment is necessary for learning in an environment with delayed consequences, and hence the learning performance may suffer.

To demonstrate the importance of Property \ref{intent} (Commitment), consider the Bipolar Chain example of Section \ref{problem}. 
Assume that the $k$th agent samples an MDP at time $t_{k, 1}$, in which the left-most edge is positive and the right-most edge is negative. 
Therefore, the $k$th agent decides to move left at $t_{k, 1}$. 
When the $k$th agent re-samples an MDP at time $t_{k, 2}$, the left-most edge may now be negative and the right-most edge may now be positive due to randomness, even if no other agent has gathered information to warrant this change in the sampled MDP. 
As a consequence, the $k$th agent moves right at $t_{k, 2}$, undoing his previous move, incurring unnecessary cost and most importantly delaying the traversal of either the left-most or the right-most edge, which would produce information valuable for all agents.

\subsubsection{Concurrent UCRL}	
At time $t_{k, m}$, the $k$th agent forms confidence bounds for the reward structure $\cR$ and the transition structure $\cP$ that define the set of statistically plausible MDPs given the posterior $\cF_{t_{k, m}}$.
The $k$th agent chooses the MDP $\cM_{k, m}$ that maximizes the achievable average reward subject to these confidence bounds. 
This algorithm is deterministic and does not suffer from the flaw of Thompson resampling.
Note, however, that if at time $t_{k, m}$ of the $m$th action of the $k$th agent and at time $t_{k', m'}$ of the $m'$th action of the $k'$th agent the posterior is the same, $\cF_{t_{k,m}} \equiv \cF_{t_{k',m'}}$, the $k$th agent and the $k'$th agent will form the same MDP. 
Therefore, the agents may not always diversify their exploration efforts.

To demonstrate the importance Property \ref{diversification} (Diversity), consider the Parallel Chains example of Section \ref{problem} and assume that the parallel chains' last edge weights, $\theta_c$, are independent. Further, assume that for any pair of chains $c$, $c'$, the prior means of $\theta_c$, $\theta_{c'}$ are the same, $\mu_{0,c} = \mu_{0, c'}$, but the prior variances of $\theta_c$, $\theta_{c'}$ differ, $\sigma_{0,c} \neq \sigma_{0, c'}$. 
Then, UCRL will direct all $K$ agents to the chain with the maximum prior variance and will not diversify exploratory effort to the other $C - 1$ chains.
As the horizon $H$ gets larger, the learning performance benefits less and less from an increased number of parallel agents and the expected mean regret per agent does not improve due to lack of diversity.

\subsection{Seed Sampling Algorithms} \label{seed sampling}

We now present the concept of \textit{seed sampling}, which offers an approach to designing efficient coordinated exploration algorithms that  
satisfy the three aforementioned properties.
The idea is that each concurrent agent independently samples a random seed, such that the mapping from seed to MDP is determined by the prevailing posterior distribution. 
Independence among seeds diversifies exploratory effort among agents (Property \ref{diversification}).
If the mapping is defined in an appropriate manner, the fact that the agent maintains a consistent seed leads to a sufficient degree of 
commitment (Property \ref{intent}), while the fact that the posterior adapts to new data allows the agent to react intelligently to new information (Property \ref{adaptation}).

In the subsections that follow, we discuss ways to define the mapping from seed and posterior distribution to sample. 
Note that the mappings we present represent special cases that apply to specific problem classes. 
The idea of seed sampling is broader and can be adapted to other problem classes, as we will explore further in Section \ref{results}.

Let $\cG_0$ be a deterministic function mapping a seed $z \sim \cZ$ to MDP $\cM_0$, such that $\cM_0 \sim \cM$.
At time $t$, the deterministic function mapping $\cG_t$ is generated based on $\cG_0$ and the history of observations $\cH_{t-1}$ available up to this time. 
At the beginning of the episode, each $k$th agent samples seed $z_k \sim \cZ$. 
At time $t_{k,m}$, the $k$th agent samples an MDP according to $\cM_{k, m} = \cG_{t_{k,m}}(z_k)$.
The intuition behind the seed sampling algorithms is that each agent forms its own sample of the MDP at each time period, which is distributed according to the posterior over $\cM$ based on all agents' observations, while the randomness injected to the agent's samples remains fixed throughout the horizon, allowing the agent to maintain the necessary commitment to action sequences that span multiple time periods.

\subsubsection{Exponential-Dirichlet Seed Sampling}
The agents are uncertain about the transition structure $\cP$ over which they hold a common Dirichlet prior $\cF^{\cP}_0$. 
The prior over the transition probabilities associated with each state-action pair $(s,a)$ is Dirichlet-distributed with parameters $\alpha_{0}(s,a,s')$, for $s' \in \cS$.  
The Dirichlet parameters are incremented upon each state transition and at time $t$, the posterior given the history of observations $\cH_{t-1}$ is $\cF^{\cP}_t$. 
At time $t$, the transition probabilities from the state-action pair $(s,a)$ is Dirichlet-distributed with parameters $\alpha_{t}(s,a,s')$, for $s' \in \cS$.
At the beginning of the episode, each $k$th agent samples $|\cS|^2 |\cA|$ sequences of independent and identically distributed seeds $z_{k, s, a, s'} = \left(z_{k, s,a, s', i}, \enskip i = 1, 2, \dots, \right)$ such that $z_{k, s,a, s', i} \sim \text{Exp}(1)$. 
The mapping from seed to transition structure is defined as 
$
\cG_t(z) := \Big\{p_{s,a}(s') =  \sum_{i=1}^{\alpha_{t}(s,a,s')} z_{s,a,s',i} \Big/ \sum_{\tilde{s} \in S } \sum_{i=1}^{\alpha_{t}(s,a,\tilde{s})} z_{s,a,\tilde{s},i}, \enskip \forall (s, a, s')$ $\in \cS \times \cA \times \cS\Big\}$. 
Then, each $p_{s,a}$ in $\cG_t(z)$ is Dirichlet distributed with parameters $\alpha_{t}(s,a)$ due to the fact that (a) if $Y_1, \dots, Y_d$ are independently distributed Gamma random variables with shape parameters $a_1, \dots, a_d$, then $X = (X_1, \dots, X_d)$ with $X_i = Y_i / \sum_{j = 1}^{d} Y_j$ is $d$-dimensional Dirichlet distributed with parameters $a_1, \dots, a_d$, and (b) any Gamma with shape parameter $a$ can be represented as the sum of $a$ Exp(1) random variables \citep{gentle2013randomnumber}. 
The transition structure of the sampled MDP of the $k$th agent at time $t_{k, m}$ is given by $\cG_{t_{k,m}}(z_k)$.

\subsubsection{Standard-Gaussian Seed Sampling} \label{standard}
The agents are uncertain about the parameters $\theta \in \Re^{|\cS||\cA|}$ of the reward structure, over which they share a common normal or lognormal prior $\cF^{\cR}_{0}$ with parameters $\mu_0$ and $\Sigma_0$. 
The posterior over $\theta$ at time $t$, given the history of observations $\cH_{t-1}$ available up to this time, is $\cF^{\cR}_t$ and is normal or lognormal with parameters $\mu_t$ and $\Sigma_t$. In either case, conjugacy properties result in simple rules for updating the posterior distribution's parameters upon each observation in $\cH_{t-1}$.
Consider Example \ref{random-graph-description} and Example \ref{parallel-chains-description} of Section \ref{problem}. 
At the beginning of the episode, each $k$th agent samples seed $z_k = \cN(0, I)$.
In the case of normal prior, as in Example \ref{parallel-chains-description} (Parallel Chains), the mapping from seed to the reward structure's parameters is defined as $\cG_t(z) := \mu_t + D_t z$, where $D_t$ is the positive definite matrix such that $D_t^T D_t = \Sigma_t$. 
Then, $\cG_t(z)$ is a multivariate normal with mean vector $\mu_t$ and covariance matrix $\Sigma_t$ \citep{gentle2009computationalstats}.
In the case of lognormal prior, as in Example \ref{random-graph-description} (Maximum Reward Path),  the mapping from seed to the reward structure's parameters is $\cG_t(z) := \exp\left(\mu_t + D_t z\right)$, where $D_t$ is defined as before.
Similarly, $\cG_t(z)$ is a multivariate lognormal with parameters $\mu_t$ and $\Sigma_t$.
The reward structure of the sampled MDP of the $k$th agent at time  $t_{k, m}$ has parameters $\hat{\theta}_{k, m} = \cG_{t_{k,m}}(z_k)$.

\subsubsection{Martingalean-Gaussian Seed Sampling} \label{martingalean}
The agents are uncertain about the parameters $\theta \in \Re^{|\cS||\cA|}$ of the reward structure, over which they share a common normal or lognormal prior $\cF^{\cR}_{0}$ with parameters $\mu_0$ and $\Sigma_0$.
Define seed $z = (\hat{\theta}_0, w)$ with distribution $\cZ$ such that $\hat{\theta}_0 \sim \cN(\mu_0, \Sigma_0)$ and $\{w_j: j = 0, 1, \dots\}$ is an IID sequence of $\cN(\mu_w, \sigma_w^2)$. 
At time $t$, the history up to this time $\cH_{t-1}$ consists of observations $\{ (s_j, a_j, s'_j, r_j), j = 1, \dots, |\cH_{t-1}|\}$.
The deterministic mapping $\cG_t$ from seed $z = (\hat{\theta}_0, w)$ to reward structure parameters is a model fit to the sample $\hat{\theta}_0$ from the prior and the observations in $\cH_{t-1}$ randomly perturbed by $w$.
In the case of normal prior, $\{w_j: j = 0, 1, \dots\}$ is an IID sequence of $\cN(0, \sigma^2)$ and $r_j | \theta \sim \cN\left(\theta_{\left(s_j, a_j\right)}, \sigma^2\right)$. 
The mapping at time $t$ from seed $z = (\hat{\theta}_0, w)$ to the reward structure's parameters is defined as 
$\cG_t(z) := \arg\min_{\rho} \Big(
(\rho - \hat{\theta}_0)^T \Sigma_0^{-1} (\rho - \hat{\theta}_0) 
+ \frac{1}{\sigma^2}\sum_{j = 1}^{\left|\cH_{t-1}\right|}(o_j^T \rho - r_j - w_j)^2\Big)$, 
where $o_j$ is the one-hot vector $|\cS||\cA| \times 1$, whose positive element corresponds to the state-action pair of the $j$th observation in $\cH_{t-1}$, $r_j$ is the reward of the $j$th observation in $\cH_{t-1}$ and $w_j$ is a component of the seed which corresponds to the perturbation of the reward of the $j$th observation in $\cH_{t-1}$.
In the case of lognormal prior, $\{w_j: j = 0, 1, \dots\}$ is an IID sequence of $\cN(- \sigma^2 /2, \sigma^2)$ and $\ln r_j | \theta \sim \cN\left(\ln \theta_{\left(s_j, a_j\right)} - \sigma^2 /2, \sigma^2\right)$. 
The mapping at time $t$ from seed $z = (\hat{\theta}_0, w)$ to the reward structure's parameters is similar as in the normal case, but instead of fitting to the rewards $r_j$, we fit to $\ln r_j$.

Consider again Example \ref{parallel-chains-description} (Parallel Chains) and Example \ref{random-graph-description} (Maximum Reward Path). 
At the beginning of the episode, each $k$th agent samples seed $z_k = (\hat{\theta}_{k, 0}, w_k)$ distributed according to $\cZ$.
At time $t_{k, m}$, the $k$th agent generates $\hat{\theta}_{k, m} = \cG_t(z_k)$, which is a model fit to his sample $\hat{\theta}_{k,0}$ from the prior and to the observations in the history $\cH_{t_{k, m}-1}$ perturbed by $w_k$,
$\hat{\theta}_{k, m} = \left(O^T O  + \sigma^2 \Sigma_0^{-1} \right)^{-1} \left(O^T (R + W^k) + \sigma^2 \Sigma_0^{-1} \hat{\theta}_{k,0} \right)$
where $O$ is the $\left|\cH_{t_{k, m}-1}\right| \times |\cS||\cA|$ matrix whose $j$th row is $o_j^T$, $R$ is the $\left|\cH_{t_{k, m}-1}\right| \times 1$ vector whose $j$th element is $r_j$ in the normal prior case and $\ln r_j$ in the lognormal prior case and $W^k$ is the $\left|\cH_{t_{k, m}-1}\right| \times 1$ vector whose $j$th element is $w_{k,j} \sim \cN(0, \sigma^2)$ in the normal prior case and is $w_{k,j} \sim \cN(-\sigma^2/2, \sigma^2)$ in the lognormal prior case.
\begin{proposition} \label{prop-posterior}
Conditioned on $\F_{t_{k, m}-1}$, $r_{k, m}$ and $s_{k, m}$, $\hat{\theta}_{k, m}$ is distributed according to the posterior of $\theta$.
\end{proposition}
\begin{proposition} \label{prop-martingale}
For each agent $k$, denote as $\cT_k = \{0, \dots, t_{k, H}\}$ and consider a probability measure $\tilde{\P}_k$ defined on $\left(\Omega, \F, (\F_t)_{t \in \cT_k}\right)$, for which $\hat{\theta}_{k, 0}$ is deterministic, $\theta$ is distributed $\cN(\hat{\theta}_{k, 0}, 2 \Sigma_0)$. Then, $\hat{\theta}_{k, m}$ is a martingale with respect to $\tilde{\P}_k$.
\end{proposition}
Proposition \ref{prop-posterior} follows from Lemma 4 of \citep{lu2017ensemble} and is core to sampling an MDP that follows the posterior distribution based on the data gathered by all agents (Property \ref{adaptation}).
Proposition \ref{prop-martingale} follows from the definitions and motivates the name of this seed sampling algorithm.

\section{Computational Results} \label{results}
In this section, we present computational results that demonstrate the robustness of seed sampling algorithms of Section \ref{seed sampling} versus the baseline algorithms of Section \ref{baseline}.
In sections \ref{bipolar-chain-results} and \ref{parallel-chains-results}, we  present two simple problems that highlight the weaknesses of concurrent UCRL and Thompson resampling and demonstrate how severely performance may suffer due to violation of any among Properties \ref{adaptation}, \ref{intent}, \ref{diversification}.
In Section \ref{random-graph-results}, we demonstrate the relative efficiency of seed sampling in a more complex problem.
 
\subsection{Bipolar Chain} \label{bipolar-chain-results}
Consider the directed graph of Figure \ref{bipolar-chain-figure} and the description of Example \ref{bipolar-chain-description} in Section \ref{problem}.
The agents' objective is to maximize the accrued reward.
However, the agents do not know whether the leftmost edge $e_L$ has weight $\theta_L = N$ or whether the rightmost edge has weight $\theta_R = N$.
The agents share a common prior that assigns equal probability $p = 0.5$ to either scenario.
When any of the $K$ agents traverses $e_L$ or $e_R$ for the first time, all $K$ agents learn the true values of  $\theta_L$, $\theta_R$. 
Denote as $T$ the time when the true MDP is revealed.
The horizon is $H = 3 N / 2$. 
The horizon is selected in such a way, so that if an agent picks the wrong direction and moves in every time period towards the wrong endpoint, if the true values of $\theta_L$, $\theta_R$ are revealed before the wrong endpoint is reached, this agent has enough time to correct the trajectory and reach the correct endpoint.
The optimal reward is $R^* = N/2$ if $\theta_L = N, \theta_R = -N$ and $R^* = N/2 + 1$ if $\theta_L = -N, \theta_R = N$ because the leftmost endpoint $v_{L} = 0$ is one vertex further from the start $v_S = N/2$ than the rightmost endpoint $v_{R} = N-1$ and requires the traversal of one more penalizing edge with weight $-1$.
We now examine how seed sampling, concurrent UCRL and Thompson resampling behave in this setting.

In seed sampling, each $k$th agent samples a seed $z_k \sim \text{Bernoulli}(p)$, which remains fixed for the entire duration of the episode. 
The mapping from seed $z_{k, m}$ to sampled MDP $\cM_{k, m}$ at time $t_{k, m} < T$ is determined by $\hat{\theta}_{L, k, m}$ and $\hat{\theta}_{R, k, m}$ which are defined as
$\hat{\theta}_{L, k, m} = N \cdot \text{sign}(z_k - 0.5)$, $\hat{\theta}_{R, k, m} = -\hat{\theta}_{L, k, m}$.
After one of the $K$ agents traverses $e_L$ or $e_R$, the sampled MDP of each $k$th agent who has not terminated is the true MDP, $\hat{\theta}_{L, k, m} = \theta_L, \quad \hat{\theta}_{R, k, m} = \theta_R$, satisfying Property \ref{adaptation}.
Note that in seed sampling, among the agents who start before the true MDP is revealed, i.e., $\{k: t_{k, 0} < T\}$, half go left and half go right in expectation, satisfying Property \ref{diversification}.
Thanks to the seed $z_{k, m}$ the sampled MDP $\cM_{k, m}$ remains fixed in every step of the $k$th agent's trajectory until the true MDP is learned. 
Therefore, all agents commit to reaching either the left or the right endpoint of the chain depending on the seed they sampled, until the correct direction is discovered by one of the agents, satisfying Property \ref{intent}.
When the correct direction is revealed, the horizon $H = 3 N / 2$ allows all agents who have picked the wrong direction but have not yet terminated to change their trajectory and eventually to reach the correct endpoint.

In concurrent UCRL, the agents are initially optimistic that they can achieve the maximum attainable reward, which is $N/2 + 1$ in the scenario that the rightmost edge (i.e., the closest one to the start) has the positive weight, $\theta_R = N$.
Each $k$th agent at time $t_{k, m} < T$ chooses an MDP that is defined as 
$\hat{\theta}_{L, k, m} = -N$, $\hat{\theta}_{R, k, m} = N$.
Therefore, all agents who start before the true MDP is revealed, i.e., $\{k: t_{k, 0} < T\}$ go right, violating Property \ref{diversification}.
Note that, in this particular example, diversification is not essential to exploration, since going towards a single direction will still reveal the true MDP. 
When the correct direction is revealed and it is not the rightmost endpoint, all agents who have not terminated change their trajectory and eventually reach the correct endpoint.
If the correct direction is the rightmost endpoint, no agent has to change trajectory.

In Thompson resampling, each $k$th agent at time $t_{k, m} < T$ samples an MDP which is defined as
$\hat{\theta}_{L, k, m} = N \cdot \text{sign}\left(\text{Bernoulli}(p) - 0.5\right)$, $\hat{\theta}_{R, k, m} = -\hat{\theta}_{L, k, m}$.
Note that for two subsequent time periods $t_{k, m} < t_{k, m+1} < T$, the sampled MDP of the $k$th agent may differ due to randomness in drawing a $\text{Bernoulli}(p)$ sample each time. 
Therefore, the $k$th agent who decided to go towards one direction at time $t_{k, m}$ may change his decision and go towards the opposite direction at time $t_{k, m+1}$, violating Property \ref{intent}.
In this setting, violation of  Property \ref{intent} is detrimental to exploration.
The horizon $H$ of each one of the $K$ agents is consumed to meaningless oscillations and it is very difficult for any agent to reach either endpoint of the chain.
The larger the number of vertices in the chain is, the more unlikely becomes for any agent to ever discover the true MDP.

\begin{figure}[h]
\centering
\includegraphics[width=0.5\columnwidth]{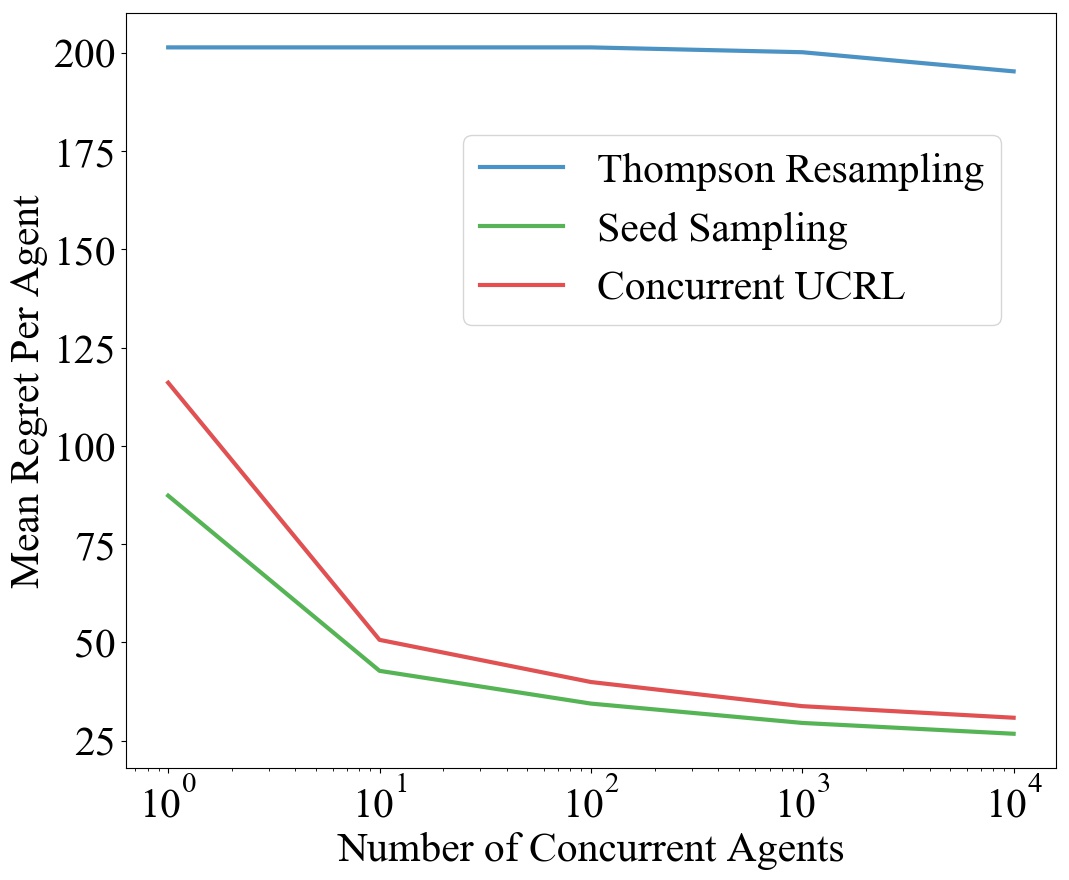}
\caption {Performance of the algorithms of Section \ref{algorithms} in the ``Bipolar Chain" example with $N = 100$ vertices and $H = 150$ horizon in terms of mean regret per agent as the number of agents increases.}
\label{bipolar-chain-mean-regret}
\end{figure}

Figure \ref{bipolar-chain-mean-regret} shows the mean regret per agent for $N = 100$ number of vertices in the chain as the number of concurrent agents increases.
The figure presents averages over hundreds of simulations.
As the number of agents increases, in seed sampling and concurrent UCRL more and more agents have not yet started their trajectory or moved further away from the start towards the wrong direction the moment the true MDP is revealed.
Therefore the mean regret per agent decreases.
On the other hand, in Thompson resampling, the lack of commitment to exploring either direction prevents the true MDP to be discovered before the horizon expires, even for a very large number of agents.
As a result the mean regret per agent does not improve.

\subsection{Parallel Chains} \label{parallel-chains-results}
Consider the directed graph of Figure \ref{parallel-chains-figure} and the description of Example \ref{parallel-chains-description} in Section \ref{problem}.
The agents' objective is to maximize the accrued reward by choosing the chain whose last edge has the largest weight $c^* = \argmax_{c}\theta_{c}$.
Recall that the weight of the last edge of chain $c$ is denoted as $\theta_c$ and $\theta = \left(\theta_1, \dots, \theta_C\right)$ 
When the last edge of chain $c$ is traversed, the reward is a noisy observation of $\theta_c$ such that $r_c | \theta \sim \cN(\theta_c, \sigma^2)$.
However, the agents do not know the true value $\theta$ and they share a common, well-specified prior on it.
Assume that all the $\theta_c$, $c = 1, \dots, C$ are independent and that the prior on the $c$th chain's last edge weight is $\cN(\mu_c, \sigma_c^2)$.
Further assume that $\forall c \in \{1, \dots, C\}$, the prior mean is the same, $\mu_c = \mu_0$, and the prior variance increases as we move from higher to lower chains, $\sigma_c^2 = \sigma_0^2 + c$.

In this setting, martingalean-Gaussian seed sampling, standard-Gaussian seed sampling and Thompson resampling are expected to have identical performance. 
Thanks to sampling the seeds independently, the martingalean-Gaussian seed sampling and standard-Gaussian seed sampling agents construct  MDPs in which different chains appear to be optimal.
This is also the case for Thompson resampling agents, who sample their MDPs independently from the prior.
As a result, the martingalean-Gaussian seed sampling, standard-Gaussian seed sampling and Thompson resampling agents are directed to different chains and satisfy Property \ref{diversification}.
Note that, unlike martingalean-Gaussian seed sampling and standard-Gaussian seed sampling, Thompson resampling does not satisfy Property \ref{intent} but in this setting this does not impact the learning performance.
A Thompson resampling agent $k$ may draw an MDP at time $t_{k, 0}$ in which chain $c$ is optimal, but due to resampling at time $t_{k, 1}$ his belief may change and another chain $c' \neq c$ may appear to be optimal.
However, since transitions from one chain to another are not possible in the directed graph of Figure \ref{parallel-chains-figure}, the agent has no choice but exploring his initial choice, which is chain $c$.
Even if inherently Thompson resampling lacks the commitment of martingalean-Gaussian seed sampling and standard-Gaussian seed sampling, the structure of the problem forces the Thompson resampling agents to maintain intent and perform equally well.

On the other hand, the concurrent UCRL agents are optimistic that they can achieve the maximum attainable reward and they are all directed to chain $C$ for which the upper confidence bound of the weight of the last edge is the largest.
Once enough agents have traversed the last edge of chain $C$ and the posterior variance on $\theta_C$ becomes lower than the prior variance on $\theta_{C-1}$, the optimistic driven behavior directs the agents who have not left the source to chain $C-1$.
As long as there are agents who have not left the source, this way of exploration repeats until some agents are directed to chain $1$.
The violation of Property \ref{diversification} leads to a wasteful allocation of the agents' exploratory effort, as all agents are directed to gather similar information.
This is detrimental to learning performance, as an agent $k$ with a later activation time $t_{k, 0}$ will not have all the information to make the optimal choice of chain that he could have made if the agents who started before him had explored all the chains.

Consider the specification of the problem with $C = 10$ chains, horizon (or equivalently number of vertices in each chain) $H = 5$, $\theta_c \sim \cN(0, 100 + c)$, $\forall c \in \{1, \dots, C\}$ and likelihood of observed reward when the last edge of chain $c$ is traversed $r_c | \theta_c \sim \cN(\theta_c, 1)$.

\begin{figure}[H]
\centering
\subfloat[{\small Mean regret per agent as the number of agents increases.}]
{\makebox[0.9\columnwidth] 
{\includegraphics[width=0.5\columnwidth]{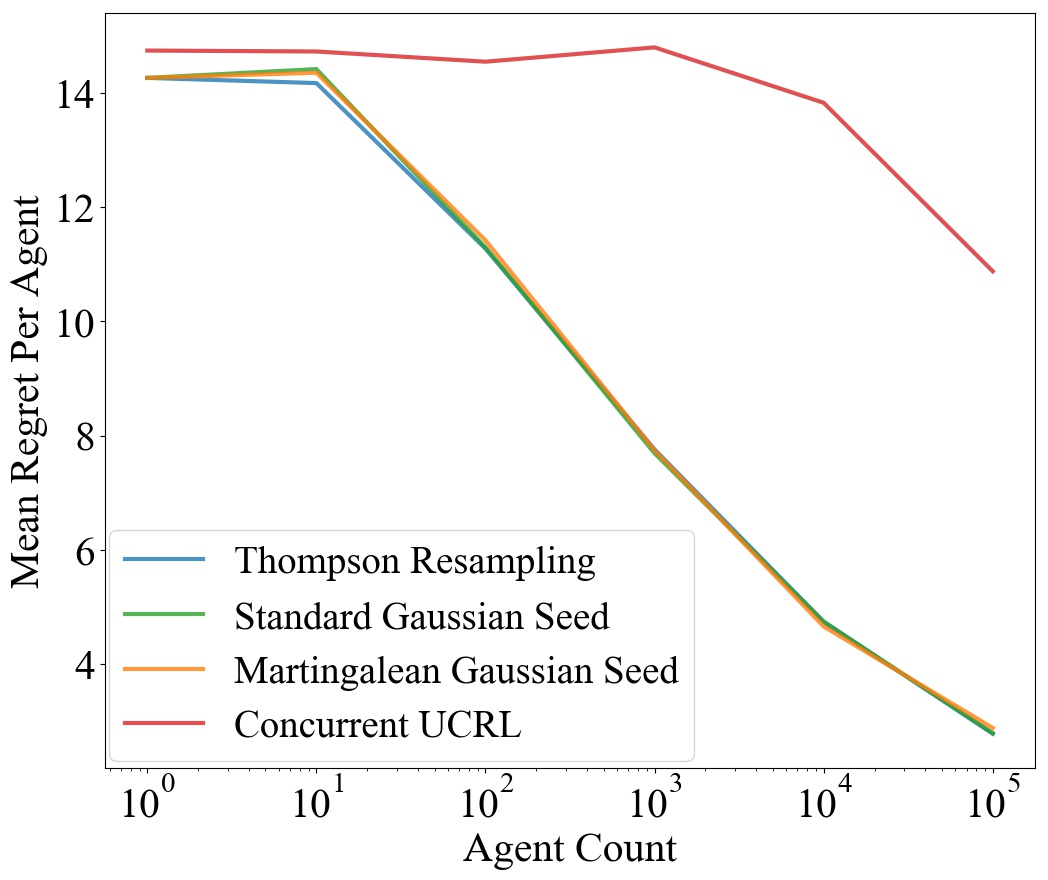}
\label{parallel-chains-mean-regret}}}
\\
\subfloat[{\small Cumulative regret of 100,000 concurrent agents ordered in ascending activation time.}]
{\makebox[0.9\columnwidth] 
{\includegraphics[width=0.5\columnwidth]{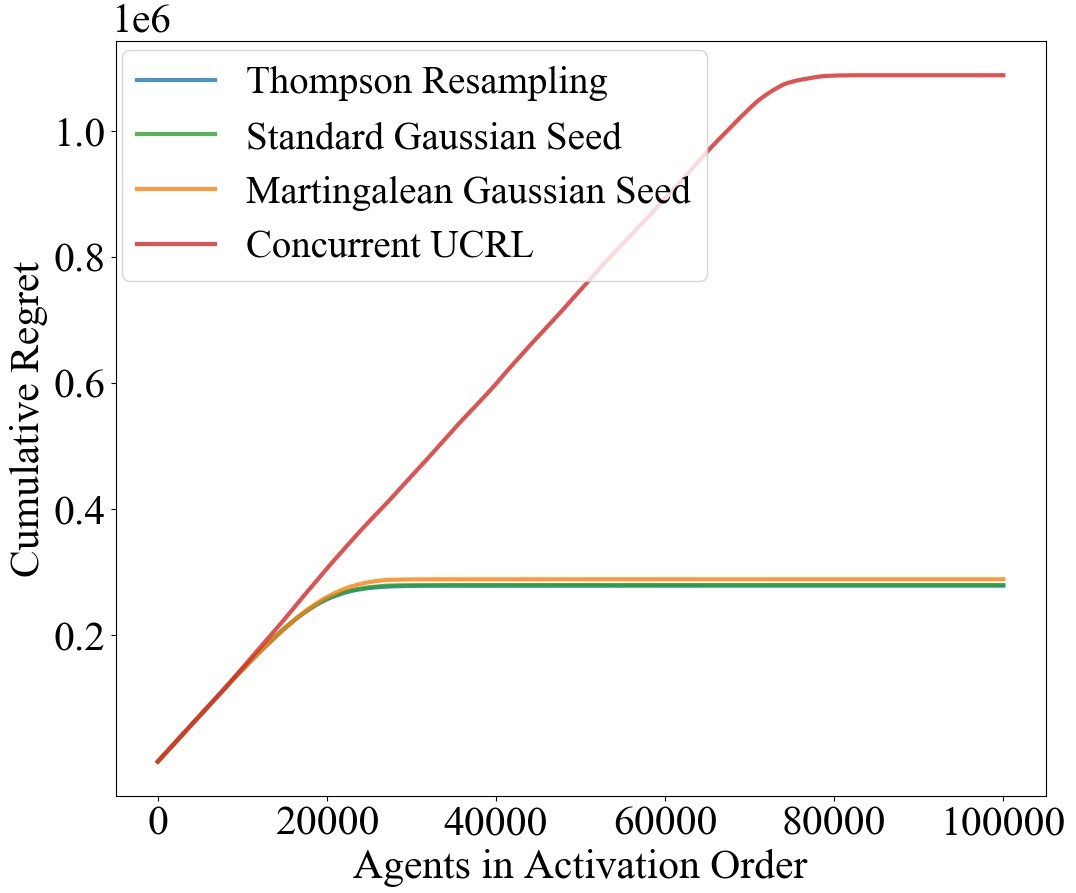}
\label{parallel-chains-cumulative-regret}}}
\caption{Performance of the algorithms of Section \ref{algorithms} in the ``Parallel Chains" example with $C = 10$ chains, $H = 5$ number of vertices per chain, $\theta_c \sim \cN(0, 100 + c)$, $\forall c \in \{1, \dots, 10\}$.}
\end{figure}

Figure \ref{parallel-chains-mean-regret} shows the mean regret per agent achieved by the algorithms as the number of agents increases and Figure \ref{parallel-chains-cumulative-regret} shows the cumulative regret of $100,000$ concurrent agents, with the agents ordered in ascending activation time $t_{k, 0}$. 
Both figures present averages over hundreds of simulations.

The figures demonstrate that UCB approaches to concurrent reinforcement learning do not efficiently coordinate, and as a result, performance may suffer severely.
In order for concurrent UCRL to achieve the same mean regret per agent that martingalean-Gaussian seed sampling, standard-Gaussian seed sampling and Thompson resampling achieve with 100 agents, 100,000 agents are required.

\subsection{Maximum Reward Path} \label{random-graph-results}
We now present the performance of the algorithms in a more complex problem. 
Consider the description of Example \ref{random-graph-description}.
The agents start from the same vertex and their goal is to make $H$ edge traversals that will return the highest reward.
Initially, the agents do not know the edge weights.
The edge weights are assumed to be independent and the agents share a common prior $\cN(\mu_e, \sigma_e^2)$ over $\ln \theta_e$, $\forall e \in \cE$.
Every time edge $e$ is traversed, the observed reward $r_e$ is distributed according to $\ln r_e | \theta_e \sim \cN(\ln \theta_e - \sigma^2/2, \sigma^2)$ and the common posterior of all agents is updated according to
$\mu_e \leftarrow \frac{\sigma^2 \mu_e + \sigma_e^2 \left(\ln r_e + \sigma^2/2\right)}{\sigma_e^2 + \sigma^2}$ and $\sigma^2_e \leftarrow \frac{\sigma_e^2 \sigma^2}{\sigma_e^2 + \sigma^2}$.
The $k$th agent, at time $t_{k, m}$, $m = 1, \dots, H$, constructs MDP $\cM_{k, m} = \hat{\theta}_{k,m} = \{\hat{\theta}_{k, m, e}, e \in \cE \}$ and from vertex $v_{k, m}$ computes the maximum reward path of $H-m+1$ steps.
 
In Thompson resampling, the $k$th agent's sampled MDPs at time $t_{k, m}$ and time $t_{k, m+1}$ may differ significantly.
Part of this difference is due to the fact that for some edges observations were made between $t_{k, m}$ and $t_{k, m+1}$.
However, $\cM_{k, m}$ and $\cM_{k, m+1}$ may also have different weights for edges that were not traversed between $t_{k, m}$ and $t_{k, m+1}$ due to randomness.
Therefore, the $k$th agent may be enticed to redirect towards an edge which has a large weight in $\cM_{k, m+1}$ but did not have a large weight in $\cM_{k, m}$, even if this change in beliefs is not substantiated by true observations.
As a result, Thompson resampling agents violate Property \ref{intent} and are susceptible to myopic behavior, which harms the agents' ability to explore deep in the graph in order to identify the maximum reward path of fixed length.

In concurrent UCRL, the agents are immune to the distractions suffered by Thompson resampling agents, as they construct deterministic upper confidence bounds from the common posteriors on the edge weights.
However, the $k$th agent at time $t_{k, m}$ and the $k'$th agent at time $t_{k', m'} > t_{k, m}$  have identical beliefs on all the edges that have not been observed between $t_{k, m}$ and $t_{k', m'}$. 
Therefore, the path chosen by $k'$th agent may be very similar or even identical to the path chosen by $k$th agent.
This lack of diversity (Property \ref{diversification}) delays the exploration of the graph's edges and the identification of the maximum reward path of fixed length.

In standard-Gaussian seed sampling or martingalean-Gaussian seed sampling, each agent samples a seed independently and constructs an MDP by using this seed and the mapping detailed in Section \ref{standard} or Section \ref{martingalean} respectively.
The fact that each agent samples a seed independently leads, thanks to randomness, to MDPs with very different edge weights for the edges that have not been traversed yet. As a result, agents pursue diverse paths.
At the same time, maintaining a constant seed ensures that each agent adjusts his beliefs in subsequent time periods in a manner that is consistent with the observations made by all agents and not driven by further randomness that would be distracting to the agent's exploration. 

Consider the specification of the problem in which we sample Erd{\H o}s-R{\' e}nyi graphs with number of vertices $N = 100$ and edge probability $p = 2 \ln N / N$.
The edge weights $\theta$ are independent and the common prior of the agents on the edge weights is $\ln \theta_e \sim \cN(0, 4)$, $\forall e \in \cE$.
When edge $e$ is traversed, the observed reward $r_e$ is distributed according to $\ln r_e | \theta_e \sim \cN(\ln \theta_e - 0.005, 0.01)$. 
The horizon (i.e. length of the maximum reward path to be found) is $H = 10$.

\begin{figure}[H]
\centering
\includegraphics[width=0.5\columnwidth]{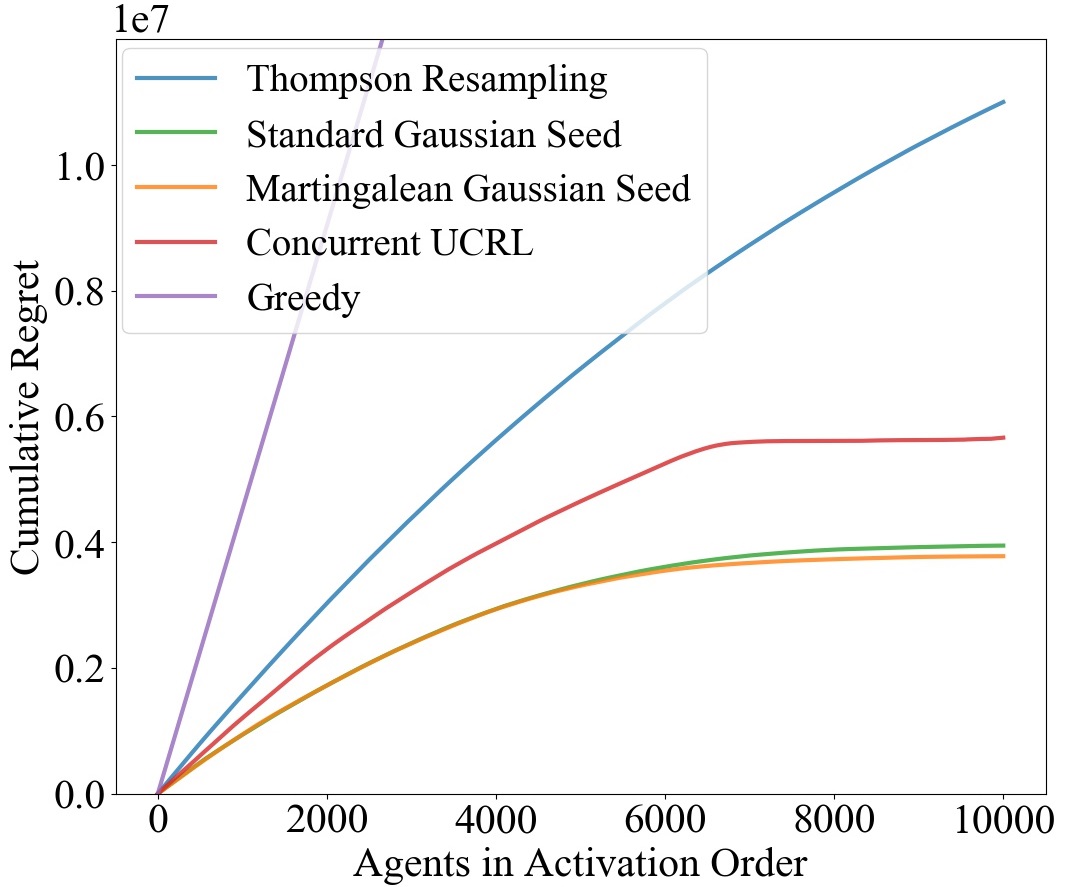}
\caption{Performance of the algorithms of Section \ref{algorithms} and of the Greedy algorithm in the ``Maximum Reward Path" example with $N = 100$ nodes, $p = 2 \ln N / N$ edge probability, $\ln \theta_e \sim \cN(0, 4)$, $\forall e \in \cE$, $H = 10$ horizon in terms of cumulative regret of 10,000 concurrent agents ordered in ascending activation time.
\label{max-cost-normal-cumulative-regret}}
\end{figure}

In Figure \ref{max-cost-normal-cumulative-regret}, we show the performance of all algorithms. The results are averaged over hundreds of simulations.
Standard-Gaussian seed sampling and martingalean-Gaussian seed sampling achieve the lowest cumulative regret, as they adhere to all the properties of coordinated exploration.
Concurrent UCRL follows with 49.9\% higher cumulative regret than the seed sampling algorithms.
Concurrent UCRL does not satisfy the diversity property, and incurs much larger cumulative regret than the seed sampling algorithms, but does better than Thompson resampling because, unlike the latter, it satisfies the commitment property which is essential in deep exploration.
Thompson resampling has 191\% higher cumulative regret than the seed sampling algorithms.

\section{Closing Remarks}
Concurrent reinforcement learning is poised to play an important role across many applications, ranging from web services to autonomous vehicles to healthcare, 
where each agent is responsible for a user, vehicle or patient.  To learn efficiently in such settings, agents should coordinate the exploratory effort.
We presented three properties that are essential to efficient coordinated exploration: real-time \textit{adaptivity} to shared observations, \textit{commitment} to carry through with action sequences that reveal new information, and \textit{diversity} across learning opportunities pursued by different agents.
We demonstrated that optimism-based approaches fall short with respect to diversity, while naive extensions of Thompson sampling lack the requisite level of commitment.
We proposed \textit{seed sampling}, a novel extension of PSRL that does satisfy these properties.  We presented several seed sampling schemes, customized for particular priors and likelihoods, but the seed sampling concept transcends these specific cases, offering a general approach to more broadly designing effective coordination algorithms for concurrent reinforcement learning.  Much work remains to be done on this topic.  For starters, it would be useful to develop a mathematical theory that sharpens understanding of the efficiency and robustness of seeding schemes.  Beyond that, work is required to develop seeding schemes that operate in conjunction with practical scalable reinforcement learning algorithms that approximate optimal value functions and policies for problems with intractable state and action spaces.

\bibliography{reference}

\begin{thebibliography}{17}
\providecommand{\natexlab}[1]{#1}
\providecommand{\url}[1]{\texttt{#1}}
\expandafter\ifx\csname urlstyle\endcsname\relax
  \providecommand{\doi}[1]{doi: #1}\else
  \providecommand{\doi}{doi: \begingroup \urlstyle{rm}\Url}\fi

\bibitem[Gentle(2009)]{gentle2009computationalstats}
James~E. Gentle.
\newblock \emph{Computational Statistics}.
\newblock Springer, 2009.

\bibitem[Gentle(2013)]{gentle2013randomnumber}
James~E. Gentle.
\newblock \emph{Random Number Generation and Monte Carlo Methods}.
\newblock Springer, 2013.

\bibitem[Guo and Brunskill(2015)]{guo2015concurrent}
Z.~Guo and E.~Brunskill.
\newblock Concurrent {PAC} {RL}.
\newblock In \emph{AAAI Conference on Artificial Intelligence}, pages
  2624--2630, 2015.

\bibitem[Jaksch et~al.(2010)Jaksch, Ortner, and Auer]{Jaksch2010}
Thomas Jaksch, Ronald Ortner, and Peter Auer.
\newblock Near-optimal regret bounds for reinforcement learning.
\newblock \emph{Journal of Machine Learning Research}, 11:\penalty0 1563--1600,
  2010.

\bibitem[Kearns and Singh(2002)]{Kearns2002}
Michael~J. Kearns and Satinder~P. Singh.
\newblock Near-optimal reinforcement learning in polynomial time.
\newblock \emph{Machine Learning}, 49\penalty0 (2-3):\penalty0 209--232, 2002.

\bibitem[Kim(2017)]{kim2017thompson}
Michael~Jong Kim.
\newblock Thompson sampling for stochastic control: The finite parameter case.
\newblock \emph{IEEE Transactions on Automatic Control}, 2017.

\bibitem[Lu and Van~Roy(2017)]{lu2017ensemble}
Xiuyuan Lu and Benjamin Van~Roy.
\newblock Ensemble sampling.
\newblock In \emph{NIPS}, 2017.

\bibitem[Osband and Van~Roy(2014{\natexlab{a}})]{osband2014model}
Ian Osband and Benjamin Van~Roy.
\newblock Model-based reinforcement learning and the eluder dimension.
\newblock In \emph{Advances in Neural Information Processing Systems}, pages
  1466--1474, 2014{\natexlab{a}}.

\bibitem[Osband and Van~Roy(2014{\natexlab{b}})]{osband2014near}
Ian Osband and Benjamin Van~Roy.
\newblock Near-optimal reinforcement learning in factored {MDP}s.
\newblock In \emph{Advances in Neural Information Processing Systems}, pages
  604--612, 2014{\natexlab{b}}.

\bibitem[Osband and Van~Roy(2017{\natexlab{a}})]{osband2017onoptimistic}
Ian Osband and Benjamin Van~Roy.
\newblock On optimistic versus randomized exploration in reinforcement
  learning.
\newblock In \emph{The Multi-disciplinary Conference on Reinforcement Learning
  and Decision Making}, 2017{\natexlab{a}}.

\bibitem[Osband and Van~Roy(2017{\natexlab{b}})]{osband2017posterior}
Ian Osband and Benjamin Van~Roy.
\newblock Why is posterior sampling better than optimism for reinforcement
  learning.
\newblock In \emph{ICML}, 2017{\natexlab{b}}.

\bibitem[Osband et~al.(2013)Osband, Russo, and Van~Roy]{Osband2013}
Ian Osband, Daniel Russo, and Benjamin Van~Roy.
\newblock ({M}ore) efficient reinforcement learning via posterior sampling.
\newblock In \emph{NIPS}, pages 3003--3011. Curran Associates, Inc., 2013.

\bibitem[Pazis and Parr(2013)]{pazis2013pac}
Jason Pazis and Ronald Parr.
\newblock {PAC} optimal exploration in continuous space {Markov} decision
  processes.
\newblock In \emph{AAAI}. Citeseer, 2013.

\bibitem[Pazis and Parr(2016)]{pazis2016pac}
Jason Pazis and Ronald Parr.
\newblock Efficient pac-optimal exploration in concurrent, continuous state
  mdps with delayed updates.
\newblock In \emph{AAAI}. Citeseer, 2016.

\bibitem[Russo et~al.(2017)Russo, Van~Roy, Kazerouni, Osband, and
  Wen]{russo2017tstutorial}
Daniel Russo, Benjamin Van~Roy, Abbas Kazerouni, Ian Osband, and Zheng Wen.
\newblock A tutorial on {Thompson} sampling.
\newblock \emph{arXiv preprint arXiv:1707.02038}, 2017.

\bibitem[Silver et~al.(2013)Silver, Newnham, Weller, and
  McFall]{silver2013concurrent}
D.~Silver, Barker Newnham, L, S.~Weller, and J.~McFall.
\newblock Concurrent reinforcement learning from customer interactions.
\newblock In \emph{Proceedings of The 30th International Conference on Machine
  Learning}, pages 924--932, 2013.

\bibitem[Strens(2000)]{Strens00}
Malcolm J.~A. Strens.
\newblock A {Bayesian} framework for reinforcement learning.
\newblock In \emph{ICML}, pages 943--950, 2000.

\end{thebibliography}
\bibliographystyle{plainnat}

\end{document}